\documentclass{article}

\usepackage[accepted]{icml2020}

\usepackage[utf8]{inputenc} 
\usepackage[T1]{fontenc}    
\usepackage{hyperref}       
\usepackage{url}            
\usepackage{booktabs}       
\usepackage{amsmath}
\usepackage{amsfonts}       
\usepackage{nicefrac}       
\usepackage{microtype}      
\usepackage{graphicx}
\usepackage{subfigure}
\usepackage{bm}

\usepackage{xcolor}
\usepackage[normalem]{ulem}

\newcommand\draft[1]{}

\DeclareMathOperator*{\argmax}{arg\,max}

\icmltitlerunning{Calibration of Model Uncertainty for Dropout Variational Inference}

\begin{document}

\twocolumn[
\icmltitle{Calibration of Model Uncertainty for Dropout Variational Inference}

\icmlsetsymbol{equal}{*}

\begin{icmlauthorlist}
\icmlauthor{Max-Heinrich Laves}{xxx}
\icmlauthor{Sontje Ihler}{xxx}
\icmlauthor{Karl-Philipp Kortmann}{xxx}
\icmlauthor{Tobias Ortmaier}{xxx}
\end{icmlauthorlist}

\icmlaffiliation{xxx}{Leibniz Universität Hannover}

\icmlcorrespondingauthor{Max-Heinrich Laves}{laves@imes.uni-hannover.de}

\icmlkeywords{Machine Learning, ICML}

\vskip 0.3in
]

\printAffiliationsAndNotice{}  

\begin{abstract}
The model uncertainty obtained by variational Bayesian inference with Monte Carlo dropout is prone to miscalibration.
In this paper, different logit scaling methods are extended to dropout variational inference to recalibrate model uncertainty.
Expected uncertainty calibration error (UCE) is presented as a metric to measure miscalibration.
The effectiveness of recalibration is evaluated on CIFAR-10/100 and SVHN for recent CNN architectures.
Experimental results show that logit scaling considerably reduce miscalibration by means of UCE.
Well-calibrated uncertainty enables reliable rejection of uncertain predictions and robust detection of out-of-distribution data.
\end{abstract}

\section{Introduction}
\label{sec:intro}

Advances in deep learning have led to high accuracy predictions for classification tasks, making deep-learning classifiers an attractive choice for safety-critical applications like autonomous driving \cite{Chen2015} or computer-aided diagnosis \cite{Esteva2017}.
However, the high accuracy of recent deep learning models in not sufficient for such applications.
In cases, where serious decisions are made upon model's predictions, it is essential to also consider the uncertainty of these predictions.
We need to know if the prediction of a model is likely to be incorrect or if invalid input data is presented to a deep model, e.\,g.\ data that is far away from the training domain or obtained from a defective sensor.
The consequences of a false decision based on an uncertain prediction can be fatal.

\begin{figure}[t]
    \centering
    \includegraphics[scale=1.0]{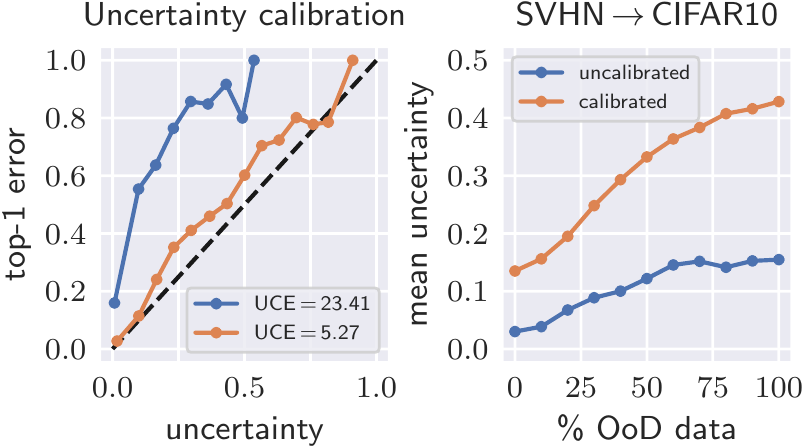}
    \caption{Calibration of uncertainty: (Left) reliability diagrams with uncertainty calibration error (UCE) and (right) detection of out-of-distribution (OoD) data. Uncalibrated uncertainty does not correspond well with the model error. Logit scaling is able to recalibrate deep Bayesian neural networks, which enables robust OoD detection. The dashed line denotes perfect calibration.}
    \label{fig:opener}
\end{figure} 

A natural expectation is that the certainty of a prediction should be directly correlated with the quality of the prediction. In other words, a prediction with a high certainty is more likely to be accurate than an uncertain prediction which is likely to be incorrect.
A common misconception is the assumption that the estimated class likelihood (of a softmax activation) can be directly used as a confidence measure for the predicted class.
This expectation is dangerous in the context of critical decision-making.
The estimated likelihood of a model trained by minimizing the negative log-likelihood (i.\,e.\ cross entropy) is highly overconfident.
That is, the estimated likelihood is considerably higher than the observed frequency of accurate predictions with that likelihood \cite{Guo2017}.

Guo et al.\ proposed calibration of the likelihood estimation by scaling the logit output of a neural network to achieve a correlation between the predicted likelihood and the expected likelihood.
However, they follow a frequentist approach, where they assume a single best point estimate of the parameters (or weights) of a neural network.
In frequentist inference, the weights of a deep model are obtained by maximum likelihood estimation \cite{Bishop2006}, and the normalized output likelihood for an unseen test input does not consider uncertainty in the weights \cite{Kendall2017}.
Weight uncertainty (also referred to as model or epistemic uncertainty) is a considerable source of predictive uncertainty for models trained on data sets of limited size \cite{Bishop2006,Kendall2017}.
Bayesian neural networks and recent advances in their approximation provide valuable mathematical tools for quantification of model uncertainty \cite{Gal2016,Kingma2013}.
Instead of assuming the existence of a single best parameter set, we place distributions over the parameters and want to consider all possible parameter configurations, weighted by their posterior.
More formally, given a training data set $ \mathcal{D} $ of labeled images and an unseen test image $ \bm{x} $ with class label $ y $, we are interested in evaluating the predictive distribution
\begin{equation}
    p(y \vert \bm{x}, \mathcal{D}) = \int p(y \vert \bm{x}, \bm{w}) p(\bm{w} \vert \mathcal{D}) \, \mathrm{d}\bm{w} ~ .
\end{equation}
This integral requires to evaluate the posterior $ p(\bm{w} \vert \mathcal{D}) $, which involves the intractable marginal likelihood \cite{Gal2016Diss}.
One practical approximation of the posterior is variational inference with Monte Carlo (MC) dropout \cite{Gal2016}.
It is commonly used to obtain epistemic uncertainty, which is caused by uncertainty in the model weights.
However, epistemic uncertainty from MC dropout still tends to be miscalibrated, i.\,e.\ the uncertainty does not correspond well with the model error \cite{Gal2017}.
The quality of uncertainty highly depends on the approximate posterior \cite{Louizos2017}.
In \cite{Lakshminarayanan2017} it is stated that MC dropout uncertainty does not allow to robustly detect out-of-distribution data.
However, calibrated uncertainty is essential as miscalibration can lead to decisions with catastrophic consequences in the aforementioned task domains.

We therefore propose a notion for perfect calibration of uncertainty and propose a definition of \emph{expected uncertainty calibration error} (UCE), derived from ECE.
We then show how current calibration techniques  (for confidence) based on logit scaling can be extended to calibrate model uncertainty. We compare calibration results for temperature scaling, vector scaling and auxiliary scaling \cite{Guo2017,Kuleshov2018} using our metric UCE as well as established ECE. We finally show how calibrated model uncertainty improves out-of-distribution (OoD) detection, as well as predictive accuracy by rejecting high-uncertainty predictions.
To the best of our knowledge, logit scaling has not been used to calibrate model uncertainty in Bayesian inference for classification.

In summary the main contributions of our work are
\begin{enumerate}
    \setlength{\itemsep}{0pt}
    \item a new metric for perfect calibration of uncertainty,
    \item derivation of logit scaling for Gaussian Dropout,
    \item first to apply logit scaling calibration to a Bayesian classifier obtained from MC Dropout, and
    \item empirical evidence that logit scaling leads to well-calibrated model uncertainty which allows robust OoD detection (in contrast to what is stated in \cite{Lakshminarayanan2017}; shown for different network architectures on CIFAR-10/100 and SVHN.
\end{enumerate}

Our code is available at: {\sloppy \href{https://github.com/link-withheld}{https://github.com/link-withheld}}.

\section{Related Work}
\label{sec:related_work}

Overconfident predictions of neural networks have been addressed by entropy regularization techniques.
Szegedy et al.\ presented label smoothing as regularization of models during supervised training for classification \cite{Szegedy2016}.
They state that a model trained with one-hot encoded labels is prone to becoming overconfident about its predictions, which causes overfitting and poor generalization.
Pereyra et al.\ link label smoothing to confidence penalty and propose a simple way to prevent overconfident networks \cite{Pereyra2017}.
Low entropy output distributions are penalized by adding the negative entropy to the training objective.
However, the referred works do not apply entropy regularization to the calibration of confidence or uncertainty.
In the last decades, several non-parametric and parametric calibration approaches such as isotonic regression \cite{Zadrozny2002} or Platt scaling \cite{Platt1999} have been presented.
Recently, temperature scaling has been demonstrated to lead to well-calibrated model likelihood in non-Bayesian deep neural networks \cite{Guo2017}.
It uses a single scalar $ T $ to scale the logits and smoothen ($ T > 1 $) or sharpen ($ T < 1 $) the softmax output and thus regularize the entropy.
Logit scaling has also been introduced to approximate categorical distributions by the Gumbel-Softmax or Concrete distribution \cite{Jang2016,Maddison2016}.
Recently, \cite{Kull2019} stated that temperature scaling does not lead to classwise-calibrated models because the single parameter $ T $ cannot calibrate each class individually.
They proposed Dirichlet calibration to address this problem.
To verify this statement, we will investigate classwise logit scaling in addition to temperature scaling.
We will show later that temperature scaling for calibrating model uncertainty in Bayesian deep learning, which takes into account all classes, does not have this shortcoming.
More complex methods, such as a neural network as auxiliary recalibration model, have been used in calibrated regression \cite{Kuleshov2018}.

\section{Methods}

In this section, we discuss how model uncertainty is obtained by Monte Carlo Gaussian dropout and how it can be calibrated with logit scaling.
We define the expected uncertainty calibration error as a new metric to quantify miscalibration and describe confidence penalty as an alternative to logit scaling.

\subsection{Uncertainty Estimation}
\label{sec:uncertainty}

We assume a general multi-class classification task with $ C $ classes.
Let input $ \bm{x} \in \mathcal{X} $ be a random variable with corresponding label $ y \in \mathcal{Y} = \{1, \ldots , C\} $.
Let $ \bm{f}_{\bm{w}}(\bm{x}) $ be the output (logits) of a neural network with weight matrices $ \bm{w} $, and with model likelihood $ p( y \! = \! c \,\vert\,  \bm{f}_{\bm{w}}(\bm{x}) ) $ for class $ c $, which is sampled from a probability vector $ \bm{p} = \bm{\sigma}_{\mathrm{SM}}(\bm{f}_{\bm{w}}(\bm{x})) $, obtained by passing the model output through the softmax function $ \bm{\sigma}_{\mathrm{SM}}(\cdot) $.
From a frequentist perspective, the softmax likelihood is often interpreted as \emph{confidence} of prediction.
Throughout this paper, we follow this definition.

To determine model \emph{uncertainty}, dropout variational inference is performed by training the model $ \bm{f}_{\bm{w}} $ with dropout \cite{Srivastava2014} and using dropout at test time to sample from the approximate posterior distribution by performing $ N $ stochastic forward passes \cite{Gal2016,Kendall2017}.
This is also referred to as MC dropout.
In MC dropout, the final probability vector is obtained by MC integration:
\begin{equation}
    \bm{p} (\bm{x}) = \frac{1}{N} \sum_{i=1}^{N} \boldsymbol{\sigma}_{\mathrm{SM}} \left( \bm{f}_{\bm{w}_{i}} (\bm{x}) \right) .
\end{equation}
The entropy of the softmax likelihood is used to describe uncertainty of prediction \cite{Kendall2017}.
In contrast to confidence as a quality measure of prediction (see §\,\ref{sec:calibration}), uncertainty takes into account the likelihoods of all $ C $ classes.
We propose to use the normalized entropy to scale the values to a range between $ 0 $ and $ 1 $:
\begin{equation}
    \tilde{\mathcal{H}}(\bm{p}) := - \frac{1}{\log C} \sum_{c=1}^{C} p^{(c)} \log p^{(c)} ~ , \quad \tilde{\mathcal{H}} \in \left[0, 1\right] .
    \label{eq:norm_entropy}
\end{equation}

Besides MC dropout there are other methods for estimating the model uncertainty such as Bayes by Backprop \cite{Blundell2015}, which uses Monte Carlo gradient estimation to learn a distribution on the weights of a neural network, or SWAG \cite{Maddox2019}, which approximates the posterior distribution with a Gaussian using the trajectory of stochastic gradient descent. These methods are however not discussed in this paper.

\subsection{Monte Carlo Gaussian Dropout}

\begin{figure}
    \centering
    \includegraphics[scale=1.0]{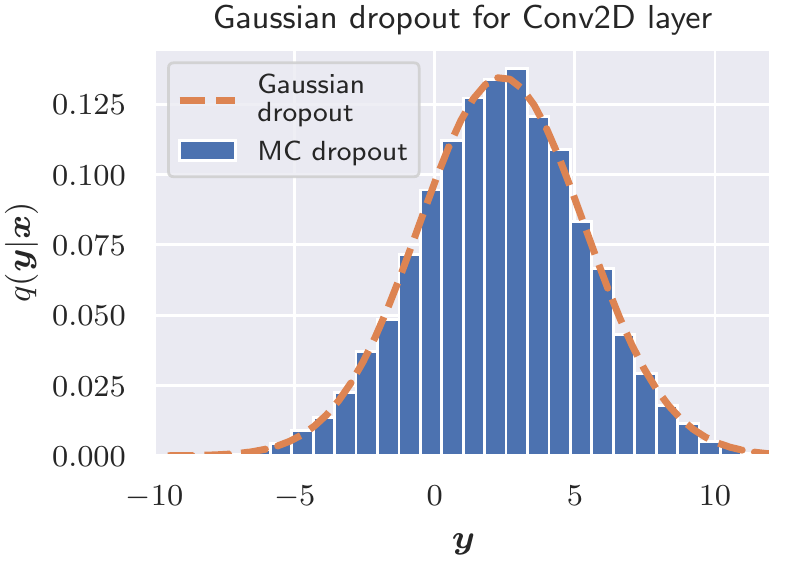}
    \caption{Implicit output distribution of MC dropout and corresponding Gaussian dropout. Gaussian dropout replaces Bernoulli dropout and allows a learnable dropout rate $ p $. The input and the weights of the convolutional layer are randomly initialized.}
    \label{fig:gaussian_dropout_clt}
\end{figure}

We will first review Gaussian dropout, which was proposed by \cite{Wang2013}, and subsequently use it to obtain model uncertainty with MC dropout.
Dropout is a stochastic regularization technique, where entries of the input $ \bm{x} $ to a weight layer $ \bm{w} $ are randomly set to zero by elementwise multiplication $ \odot $ with
\begin{gather}
    \bm{d} ~ \mathrm{where} ~ d_{j} \sim \mathsf{Bernoulli}(1-p) ~ , \\
    \bm{y} = \bm{w}^{T}(\bm{d} \odot (\bm{x} / (1-p) ) ) ~ ,
\end{gather}
with dropout rate $ p $.
This introduces Bernoulli noise during optimization and reduces overfitting of the training data.
The resulting output $ \bm{y} $ of a layer with dropout is a weighted sum of Bernoulli random variables.
Then, the central limit theorem states, that $ \bm{y} $ is approximately normally distributed (see Fig.\,\ref{fig:gaussian_dropout_clt}).
Instead of sampling from the weights and computing the resulting output, we can directly sample from the implicit Gaussian distribution of dropout
\begin{equation}
    \bm{y} \sim q( \bm{y} \vert \bm{x}) = \mathcal{N}(\mu_{\bm{y}}, \sigma_{\bm{y}}^{2})
\end{equation}
with
\begin{gather}
    \mu_{\bm{y}} = \mathbb{E}[ y_{k} ] = \sum_{j} w_{j,k} x_{j} ~ , \\
    \sigma_{\bm{y}}^{2} = \mathrm{Var}[ y_{k} ] = p/(1-p) \sum_{j} w_{j,k}^{2} x_{j}^{2} ~ ,
\end{gather}
using the reparameterization trick \cite{Kingma2015}
\begin{equation}
    y_{j} = \mu_{j} + \sigma_{j} \varepsilon_{j} ~ \mathrm{with} ~ \varepsilon_{j} \sim \mathcal{N}(0, 1) ~ .
\end{equation}
Gaussian dropout is a continuous approximation to Bernoulli dropout, and in comparison it will better approximate the true posterior distribution and is expected to provide improved uncertainty estimates \cite{Louizos2017}.
Throughout this paper, Gaussian dropout is used as a substitute to Bernoulli dropout to obtain epistemic uncertainty under the MC dropout framework.
It can efficiently be implemented in four lines of PyTorch code (see Fig.\,\ref{fig:gaussian_dropout_code}).
The dropout rate $ p $ is now a learnable parameter and does not need to be chosen carefully by hand.
In fact, $ p $ could be optimized w.r.t.\ uncertainty calibration, scaling the variance of the implicit Gaussian of dropout.
A similar approach was presented by \cite{Gal2017} using the Concrete distribution.
However, we will focus on logit scaling methods for calibration and therefore fixed $ p $ in our subsequent experiments.

Gaussian dropout has been used in the context of uncertainty estimation in prior work.
In \cite{Louizos2017}, it is used together with multiplicative normalizing flows to improve the approximate posterior.
A similar Gaussian approximation of Batch Normalization was presented in \cite{Teye2018}, where Monte Carlo Batch Normalization is proposed as approximate Bayesian inference.

\begin{figure}
    \centering
    \fbox{\begin{minipage}{0.96\columnwidth}
    \scriptsize\ttfamily
    \textbf{def} Gaussian\_dropout(x, p, layer):\\
    \hphantom{~~~~}mu = conv2d(x, layer.weight.data)\\
    \hphantom{~~~~}sigma = conv2d(x**2, layer.weight.data**2)\\
    \hphantom{~~~~}sigma = (p / (1 - p) * sigma).sqrt()\\
    \hphantom{~~~~}eps = randn\_like(mu)\\
    \hphantom{~~~~}\textbf{return} mu + sigma * eps
    \end{minipage}}
    \caption{PyTorch implementation of Gaussian dropout for a 2D convolutional layer. Gaussian dropout can be used for all common weight layers.}
    \label{fig:gaussian_dropout_code}
\end{figure}

\subsection{Calibration of Uncertainty}
\label{sec:calibration}

To give an insight into our general approach to calibration of uncertainty, we will first revisit the definition of perfect calibration of confidence \cite{Guo2017} and show how this concept can be extended to calibration of uncertainty. 

Let $ \hat{y} = \argmax \bm{p} $ be the most likely class prediction of input $ \bm{x} $ with likelihood $ \hat{p} = \max \bm{p} $ and true label $ y $. Then, following \cite{Guo2017}, \emph{perfect calibration of confidence} is defined as
\begin{equation}
    \mathbb{P} \left( \hat{y} = y \,\vert\, \hat{p} = q \right) = q , \quad \forall q \in \left[ 0, 1 \right] .
    \label{eq:perfect_calibration}
\end{equation}
That is, the probability of a correct prediction $ \hat{y} = y $ given the prediction confidence $ \hat{p} $ should exactly correspond to the prediction confidence.

From Eq.\,(\ref{eq:perfect_calibration}) and Eq.\,(\ref{eq:norm_entropy}), we define \emph{perfect calibration of uncertainty} as
\begin{equation}
    \mathbb{P} ( \hat{y} \neq y \,\vert\, \tilde{\mathcal{H}}( \bm{p} ) = q ) = q , \quad \forall q \in \left[0, 1\right] .
\end{equation}
That is, in a batch of inputs that are all predicted with uncertainty of e.\,g.\ $ 0.2 $, a top-1 error of $ 20\,\% $ is expected.
The confidence is interpreted as the probability of belonging to a particular class, which should naturally correlate with the model error of that class.
This characteristic does not generally apply to entropy, and therefore the question arises why entropy should resonate with the model error.
However, entropy is considered a measure of uncertainty, and we expect that a prediction with lower uncertainty is less likely to be false and vice versa.
In fact, our experimental results for uncalibrated models show that the confidence is as miscalibrated as the normalized entropy (see Fig.\,\ref{fig:reliability}).

\subsection{Expected Uncertainty Calibration Error (UCE)}

Due to optimizing the weights $ \bm{w} $ via minimization of the negative log-likelihood of $ p( y \,\vert\, \bm{f}_{\bm{w}}(\bm{x}) ) $, modern deep models are prone to overly confident predictions and are therefore miscalibrated \cite{Guo2017,Gal2017}.
A popular way to quantify miscalibration of neural networks with a scalar value is the expectation of the difference between predicted softmax likelihood $ \hat{p} $ and accuracy
\begin{equation}
    \mathbb{E}_{\hat{p}}\left[ \, \left| \mathbb{P} \left( \hat{y} = y \,\vert\, \hat{p} = q \right) - q \right| \, \right], \quad \forall q \in \left[ 0, 1 \right] ,
    \label{eq:ece}
\end{equation}
based on the natural expectation that confidence should linearly correlate to the likelihood of a correct prediction. This expectation of the difference can be approximated by the Expected Calibration Error (ECE) \cite{Naeini2015,Guo2017}.
The output of a neural network is partitioned into $ M $ bins with equal width and a weighted average of the difference between accuracy and confidence (softmax likelihood) is taken:
\begin{equation}
    \mathrm{ECE} = \sum_{m=1}^{M} \frac{\left| B_{m} \right|}{n} \, \big| \mathrm{acc}(B_{m}) - \mathrm{conf}(B_{m}) \big| ~ ,
\end{equation}
with total number of inputs $ n $ and set of indices $ B_{m} $ of inputs whose confidence falls into that bin (see \cite{Guo2017} for more details).
We propose the following slightly modified notion of Eq.\,(\ref{eq:ece}) to quantify miscalibration of uncertainty:
\begin{equation}
    \mathbb{E}_{\tilde{\mathcal{H}}} [ \, \vert \mathbb{P} ( \hat{y} \neq y \,\vert\, \tilde{\mathcal{H}}( \bm{p} ) = q ) - q \vert  \, ], \quad \forall q \in \left[ 0, 1 \right] .
\end{equation}
We refer to this as Expected Uncertainty Calibration Error (UCE) and approximate analogously with
\begin{equation}
    \mathrm{UCE} := \sum_{m=1}^{M} \frac{\left| B_{m} \right|}{n} \big| \mathrm{err}(B_{m}) - \mathrm{uncert}(B_{m}) \big| ~ .
    \label{eq:uce}
\end{equation}
The error per bin is defined as
\begin{equation}
    \mathrm{err}(B_{m}) := \frac{1}{\left| B_{m} \right| } \sum_{i \in B_{m}} \bm{1} (\hat{y}_{i} \neq y) ~ ,
\end{equation}
where $ \bm{1} (\hat{y}_{i} \neq y) = 1 $ and $ \bm{1} (\hat{y}_{i} = y) = 0 $.
Uncertainty per bin is defined as
\begin{equation}
    \mathrm{uncert}(B_{m}) := \frac{1}{\left| B_{m} \right| } \sum_{i \in B_{m}} \tilde{\mathcal{H}} (\bm{p}_{i}) ~ .
\end{equation}
In \cite{Kull2019}, it is stated that the ECE has a fundamental limitation.
Due to binning across all classes, over-confidence on one class can be compensated by under-confidence on another class.
Thus, a model can achieve low ECE values even if the confidence for each classes is either over- or underestimated.
They propose the classwise ECE (cECE) and, following that, we additionally define the classwise UCE (cUCE) as
\begin{equation}
    \mathrm{cUCE} := \frac{1}{C} \sum_{c=1}^{C} \mathrm{UCE}(c)
    \label{eq:cuce}
\end{equation}
to evaluate classwise calibration.
It is defined as the mean of all UCEs per class, which are denoted by $ \mathrm{UCE}(c) $.
Additionally, we plot $ \mathrm{err}(B_{m}) $ vs.\ $ \mathrm{uncert}(B_{m}) $ to create reliability diagrams and visualize calibration.

\subsection{Temperature Scaling for Dropout Variational Inference}

State-of-the-art deep neural networks are generally miscalibrated with regard to softmax likelihood \cite{Guo2017}.
However, when obtaining model uncertainty with dropout variational inference, this also tends to be not well-calibrated \cite{Louizos2017,Gal2017,Lakshminarayanan2017}.
Fig.\,\ref{fig:opener} (left) shows reliability diagrams \cite{Niculescu2005} for ResNet-101 trained on CIFAR-100.
The divergence from the identity function reveals miscalibration.
Furthermore, it is not possible to robustly detect OoD data from uncalibrated uncertainty (see Fig.\,\ref{fig:opener} (right)).
If the fraction of OoD data in a batch of test images is $ > 50\,\% $, there is almost no increase in mean uncertainty.
We first address the problem using temperature scaling, which is the most straightforward logit scaling method for recalibration.

Temperature scaling with MC dropout variational inference is derived by closely following the derivation of frequentist temperature scaling in the appendix of \cite{Guo2017}.
Let $ \left\{ \bm{z}_{1,j}, \ldots , \bm{z}_{N,j} \right\} $ be a set of logit vectors obtained by MC dropout with $ N $ stochastic forward passes for each input $ \bm{x}_{j} \in \left\{ \bm{x}_{1}, \ldots , \bm{x}_{M} \right\} $
with true labels $ \left\{ y_{1}, \ldots , y_{M} \right\} $.
Temperature scaling is the solution $ \hat{p} $ to entropy maximization
\begin{equation}
    \underset{\hat{p}}{\max} ~ - \frac{1}{N} \sum_{i=1}^{N} \sum_{j=1}^{M} \sum_{c=1}^{C} \hat{p} \left( \bm{z}_{i,j} \right)^{(c)} \log \hat{p} \left( \bm{z}_{i,j} \right)^{(c)} ,
\end{equation}
subject to
\begin{equation}
    \hat{p} (\bm{z}_{i,j})^{(c)} \geq 0 \quad \forall i,j,c ~ ,
\end{equation}
\begin{equation}
    \sum_{c=1}^{C} \hat{p} (\bm{z}_{j})^{(c)} = 1 \quad \forall j ~ ,
\end{equation}
\begin{equation}
    \frac{1}{N} \sum_{i=1}^{N} \sum_{j=1}^{M} z_{i,j}^{(y_{j})} = \frac{1}{N} \sum_{i=1}^{N} \sum_{j=1}^{M} \sum_{c=1}^{C} z_{i,j}^{(c)} \hat{p} ( \bm{z}_{i,j})^{(c)} .
\end{equation}
Guo et\,al.\ solve this constrained optimization problem with the method of Lagrange multipliers.
We skip reviewing their proof as one can see that the solution to $ \hat{p} $ in the case of MC dropout integration provides
\begin{align}
    \frac{1}{N} \sum_{i=1}^{N} \hat{p}_{i} \left( \bm{z}_{j} \right)^{(c)} &= \frac{1}{N} \sum_{i=1}^{N} \frac{e^{\lambda z_{i,j}^{(c)}}}{\sum_{\ell=1}^{C} e^{\lambda z_{i,j}^{(\ell)} }} \\
    &= \frac{1}{N} \sum_{i=1}^{N} \boldsymbol{\sigma}_{\mathrm{SM}} \left( \lambda \bm{f}_{\bm{w}_{i}} ( \bm{x}_{j} )\right)^{(c)} ,
\end{align}
which yields temperature scaling for $ \lambda = T^{-1} $ \cite{Guo2017}.
A scalar parameter cannot rescale the class logits individually.
Thus, more complex logit scaling can be derived by using any function at this point to smoothen or sharpen the softmax output (see next section).

In this work, Gaussian dropout is inserted between each weight layer with fixed dropout rate of $ p = 0.2 $.
Temperature scaling with $ T > 0 $ is inserted before final softmax activation and before MC integration:
\begin{equation}
    \hat{\bm{p}} (\bm{x}) = \frac{1}{N} \sum_{i=1}^{N} \boldsymbol{\sigma}_{\mathrm{SM}} \left( T^{-1} \bm{f}_{\bm{w}_{i}} (\bm{x}) \right) .
\end{equation}
First, $ \bm{f}_{\bm{w}} $ is trained with Gaussian dropout until convergence on the training set.
Next, we fix the parameters $ \bm{w} $ and optimize $ T $ with respect to the negative log-likelihood on a separate calibration set using MC Gaussian dropout.
This is equivalent to maximizing the entropy of $ \hat{\bm{p}} $ \cite{Guo2017}.

\subsection{Classwise Logit Scaling}
\label{sec:aux_method}

It is stated by \cite{Kull2019} that temperature scaling would be inferior to more complex calibration methods when compared by means of classwise calibration.
In \cite{Guo2017}, temperature scaling is used to calibrate the confidence that takes into account only one class probability.
In contrast, we use temperature scaling to calibrate the model uncertainty, expressed via normalized entropy.
This considers all class probabilities and thus, we hypothesize that temperature scaling implicitly leads to well-calibrated classwise uncertainty.

To demonstrate this experimentally, we implement vector scaling and auxiliary scaling and compare them using classwise UCE.
\emph{Vector scaling} is a multi-class extension of temperature scaling, where an individual scaling factor for each class is used to scale the final softmax output:
\begin{equation}
    \hat{\bm{p}}_{i} (\bm{x}) = \boldsymbol{\sigma}_{\mathrm{SM}} \left( \bm{T} \bm{f}_{\bm{w}_{i}} (\bm{x}) \right) ~ ,
\end{equation}
with $ \bm{T} = \mathrm{diag} (t_{1}, \ldots, t_{C}) $.
\emph{Auxiliary scaling} makes use of a more powerful auxiliary recalibration model $ \bm{R}_{\bm{\theta}} $ consisting of a two-layer fully-connected network with $ C $ hidden units and leaky ReLU activations after the hidden layer:
\begin{equation}
    \hat{\bm{p}}_{i} (\bm{x}) = \boldsymbol{\sigma}_{\mathrm{SM}} \left( \bm{R}_{\bm{\theta}} ( \bm{f}_{\bm{w}_{i}} (\bm{x}) ) \right) ~ ,
\end{equation}
which is inspired by \cite{Kuleshov2018}.
The intuition behind this is that recalibration may require a more complex function than simple scaling.
Both $ \bm{T} $ and the parameters $ \bm{\theta} $ of the auxiliary model are optimized w.r.t.\ negative log-likelihood in a separate calibration phase by gradient descent. 
We initialize with $ t_{j} \leftarrow 1 $ and $ \bm{\theta}_{1,2} \leftarrow \bm{I}_{C} $, respectively.
Thus, recalibration is started form the identity function.

It must be emphasized that in contrast to temperature scaling, both vector and aux scaling can change the maximum of the softmax and thus affect model accuracy.

\subsection{Confidence Penalty}
\label{sec:conf_penalty}

Additionally, we compare temperature scaling to entropy regularization, where low entropy output distributions are penalized by adding the negative entropy $ \mathcal{H} $ of the softmax output to the negative log-likelihood training objective, weighted by an additional hyperparameter $ \beta $.
This leads to the following optimization function:
\begin{equation}
    \mathcal{L}_{\mathrm{CP}}(\bm{w}) = - \sum_{\mathcal{X}, \mathcal{Y}} \log \bm{p}_{\bm{w}} (\bm{y} \vert \bm{x}) - \beta \, \mathcal{H} \left( \bm{p}_{\bm{w}}(\bm{y} \vert \bm{x}) \right) ~ .
    \label{eq:conf_penalty}
\end{equation}
We reproduce the experiment of Pereyra et al.\ on supervised image classification \cite{Pereyra2017} and compare the quality of calibration of confidence and uncertainty to logit scaling calibration methods.
Calibration by confidence penalty must be performed during the training and cannot be done afterwards.
Thus, a separate calibration phase is omitted.

\section{Experiments}
\label{sec:experiments}

The experimental results are presented threefold:
First, the proposed logit scaling methods are used to calibrate confidence and uncertainty and are compared with entropy regulation; second, predictions with high uncertainty are rejected; and third, the effect of out-of-distribution data on uncertainty is analyzed.
All models were trained from random initialization.
More details on the training procedure can be found in the appendix.

\subsection{Uncertainty Calibration}
\label{sec:exp_uncert}

To show the effectiveness of uncertainty calibration, we train ResNet-34 \cite{He2016} and DenseNet-121 \cite{Huang2017} on CIFAR-10 \cite{Krizhevsky2009} and SVHN \cite{SVHN}, as well as ResNet-101 and DenseNet-169 on CIFAR-100 with Gaussian dropout until convergence.
We mainly focus on the calibration of uncertainty obtained by performing $ N=25 $ forward passes with MC Gaussian dropout.
Additionally, we reproduce the experiments of \cite{Guo2017} and analyze calibration of frequentist confidence $ \hat{p} = \max \bm{p} $ along with likelihood values $ \hat{p} = \max N^{-1} \sum_{i=1}^{N} \bm{p}_{i} $ from MC dropout.
Subsequently, the models are calibrated using the previously mentioned logit scaling methods.
The validation set with 5,000 images is used as calibration set.
We additionally train all networks in the exact same manner with confidence penalty loss with fixed $ \beta = 0.1 $.
The proposed UCE and classwise UCE metrics are used to quantify calibration of uncertainty.
Reliability diagrams (top-1 error vs.\ uncertainty) are used to visualize (mis-)calibration.
Classwise UCE values are given in Tab.\,\ref{tab:results} and the reliability diagrams show the corresponding UCE.

\subsection{Rejection of Uncertain Predictions}

An example application of well-calibrated uncertainty is the rejection of uncertain predictions.
In e.\,g.\ a medical imaging scenario, a critical decision should only be made on the basis of reliable predictions.
We define an uncertainty threshold $ \mathcal{H}_{\mathrm{max}} $ and reject all predictions from the test set where $ \tilde{\mathcal{H}}(\bm{p}) > \mathcal{H}_{\mathrm{max}} $.
A decrease in false predictions of the remaining test set is expected.

\subsection{Out-of-Distribution Detection}

Deep neural networks only provide reliable predictions for data on which they have been trained.
In practice, however, the trained network will encounter samples that lie outside the distribution of training data.
Problematically, a miscalibrated model will still produce highly confident estimates for such out-of-distribution (OoD) data \cite{Lee2018}.

To our surprise, Bayesian neural networks have not been extensively studied for out-of-distribution detection.
Epistemic uncertainty from MC dropout was successfully used to detect OoD samples in neural machine translation \cite{Xiao2019}.
We reproduce the experiments presented by \cite{Lakshminarayanan2017}, where predictive uncertainty obtained from deep ensembles is used to detect if data from CIFAR10 is provided to a network trained on SVHN.
They state that uncertainty produced by MC dropout is over-confident and cannot robustly detect OoD data.
We expect that well-calibrated uncertainty from Bayesian methods allows us to detect if data from CIFAR10 is presented to a deep model trained on SVHN.
However, the SVHN data set shows house numbers and the CIFAR data set contains everyday objects and animals; the data domains are overly disjoint.
To demonstrate the OoD detection ability under more difficult conditions, we additionally provide images from CIFAR100 to a deep model trained on CIFAR10 (note that both CIFAR data sets have no mutual classes).

In this experiment, we compose a batch of 100 images from the test set of the training domain and stepwise replace images with out-of-distribution data.
In practice, it is expected that models are applied to a mix of known and unknown classes.
After each step, we evaluate the batch mean uncertainty and expect, that the mean uncertainty increases as a function of the fraction of OoD data.

\section{Results}
\label{sec:results}

\begin{table*}
    \small
    \centering
    \caption{Classwise ECE and UCE test set results in \% ($ M=15 $ bins). 0\,\% means perfect calibration.}
    \begin{tabular}{cccccccccccc}
        \toprule
         & & \multicolumn{2}{c}{uncalibrated} & \multicolumn{2}{c}{conf. penalty} & \multicolumn{2}{c}{temp. scaling} & \multicolumn{2}{c}{vector scaling} & \multicolumn{2}{c}{aux. scaling} \\
        \cmidrule(lr){3-4} \cmidrule(lr){5-6} \cmidrule(lr){7-8} \cmidrule(lr){9-10} \cmidrule(lr){11-12}
        Data Set & Model & cECE & cUCE & cECE & cUCE & cECE & cUCE & cECE & cUCE & cECE & cUCE \\
        \midrule
        CIFAR-10 & ResNet-34     & 4.46 & 4.03 & 8.29 & 19.8 & \textbf{1.95} & 3.68 & 2.09 & 3.73 & 2.10 & \textbf{2.38} \\
        CIFAR-10 & DenseNet-121   & 10.1 & 9.52 & 8.49 & 18.5 & 3.05 & 5.72 & 3.15 & 6.09 & \textbf{2.98} & \textbf{4.55} \\
        CIFAR-100 & ResNet-101   & 20.5 & 23.2 & 14.6 & 19.4 & 10.8 & 11.5 & \textbf{10.7} & \textbf{11.4} & 32.9 & 35.3 \\
        CIFAR-100 & DenseNet-169 & 32.4 & 37.1 & 15.6 & 20.6 & 12.9 & 13.9 & \textbf{12.8} & \textbf{13.8} & 48.9 & 52.6 \\
        SVHN & ResNet-34         & 2.37 & 2.07 & 9.11 & 22.3 & 1.47 & 3.47 & 1.44 & 3.43 & \textbf{1.34} & \textbf{1.85} \\
        SVHN & DenseNet-121      & 2.91 & 2.47 & 7.53 & 19.7 & 2.06 & 5.08 & 1.96 & 4.88 & \textbf{1.51} & \textbf{2.46} \\
        \bottomrule
    \end{tabular}
    \label{tab:results}
\end{table*}
\begin{figure*}
    \centering
    \begin{tabular}{cc}
        \small ResNet-101/CIFAR100 & \small DenseNet-169/CIFAR100 \\
        \includegraphics[width=0.98\columnwidth]{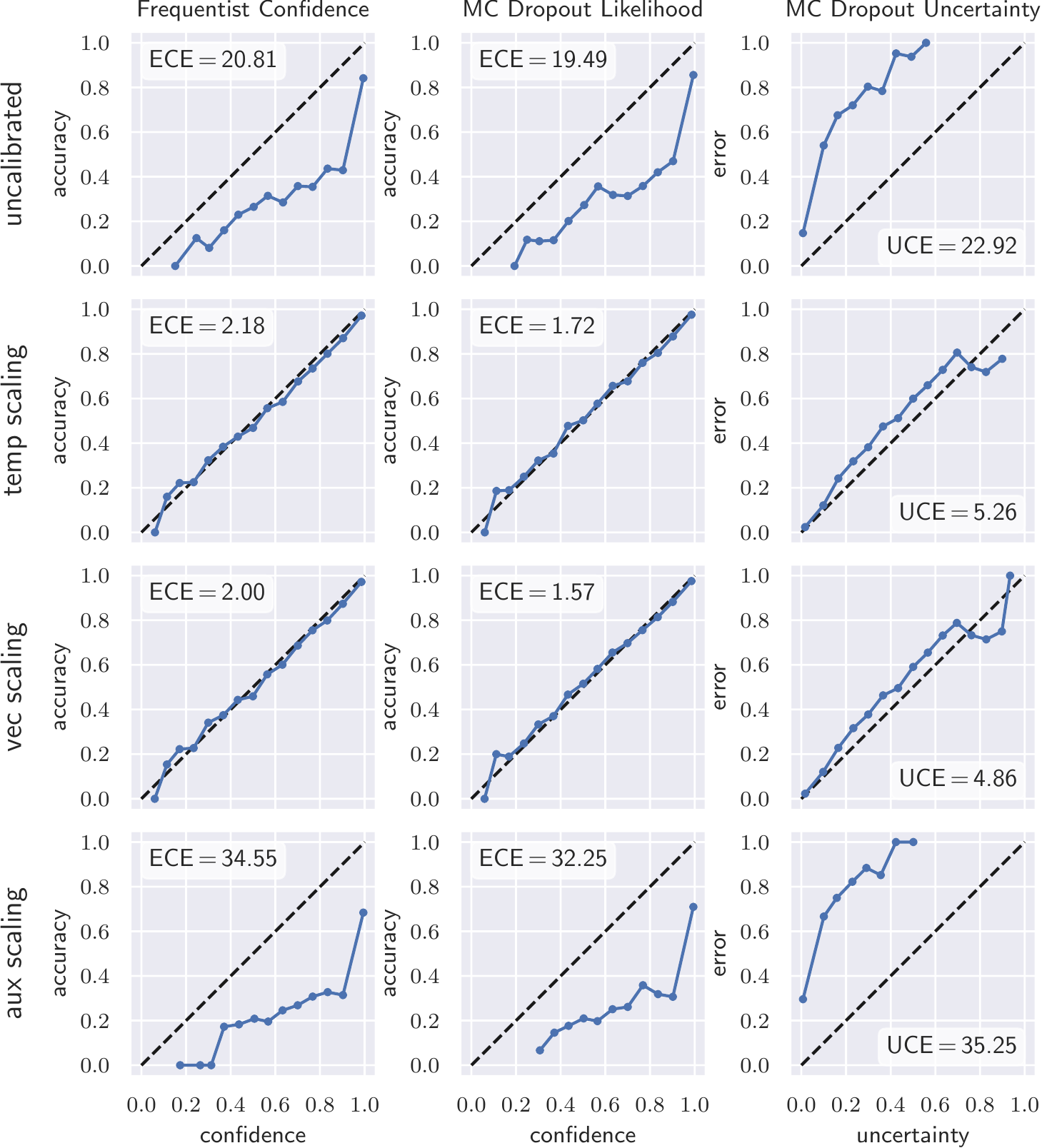} & \includegraphics[width=0.98\columnwidth]{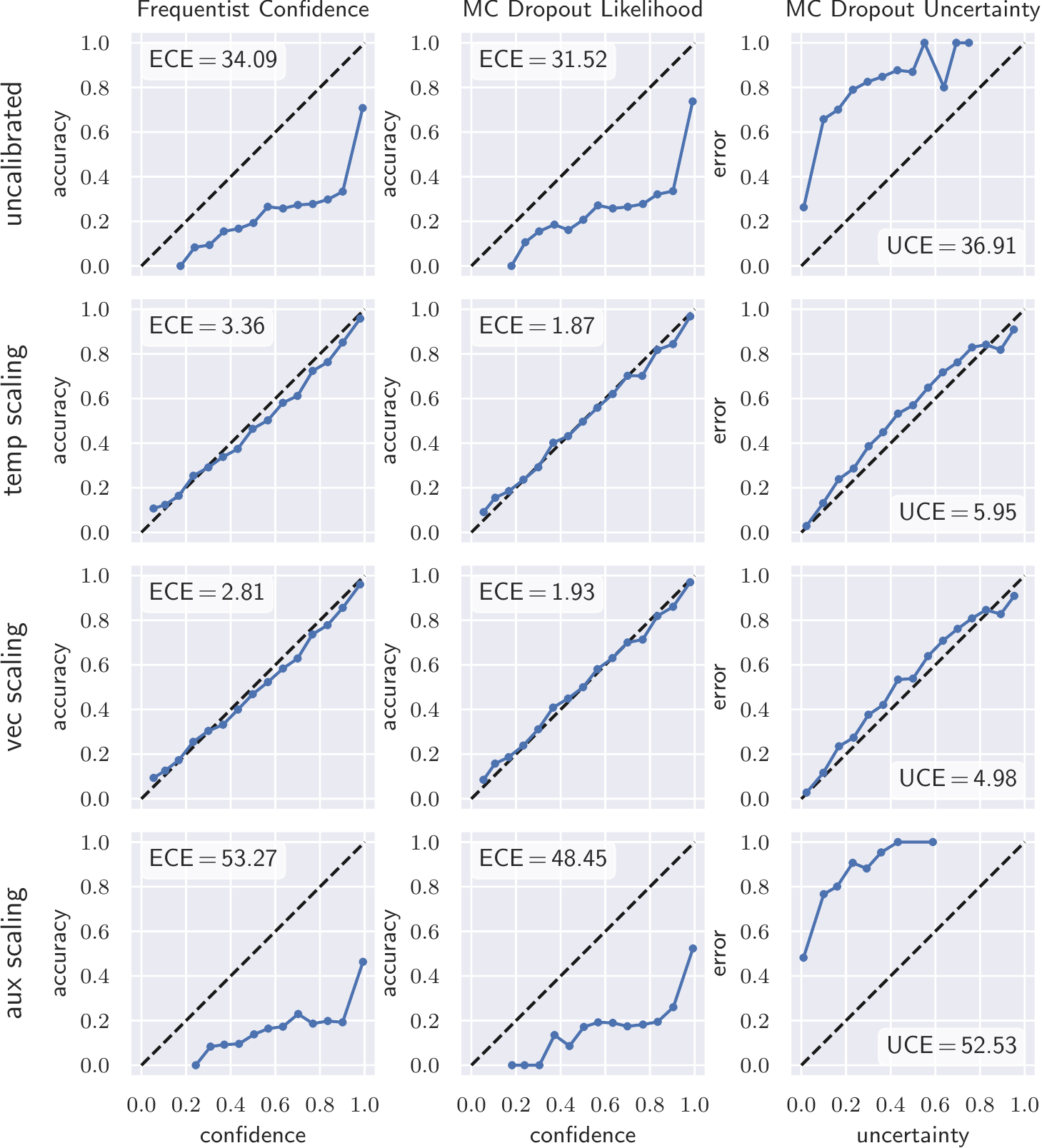}
    \end{tabular}
    \caption{Reliability diagrams ($ M = 15 $ bins) on CIFAR-100 for ResNet-101 (left) and DenseNet-169 (right). Top row: Uncalibrated frequentist confidence, and likelihood and uncertainty obtained by MC Gaussian dropout. The following rows show the results of the logit scaling methods. The dotted lines illustrates perfect calibration. Additional diagrams can be found in the supplemental material.}
    \label{fig:reliability}
\end{figure*}
\begin{figure*}
    \centering
    \begin{tabular}{cc}
    \small Rejection of unreliable predictions & \small Out-of-Distribution Detection \\
    \includegraphics[width=0.98\columnwidth]{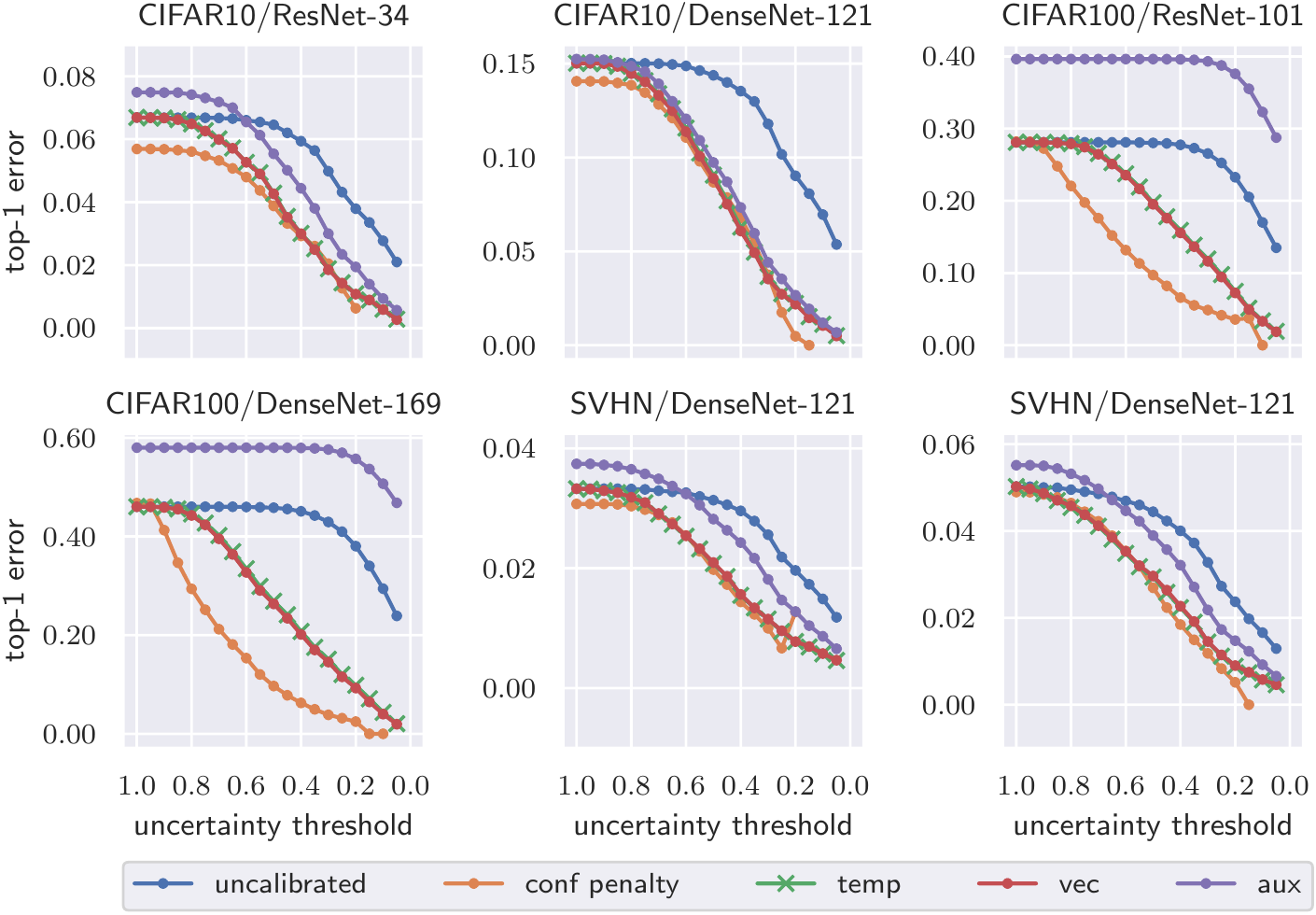} & \includegraphics[width=0.98\columnwidth]{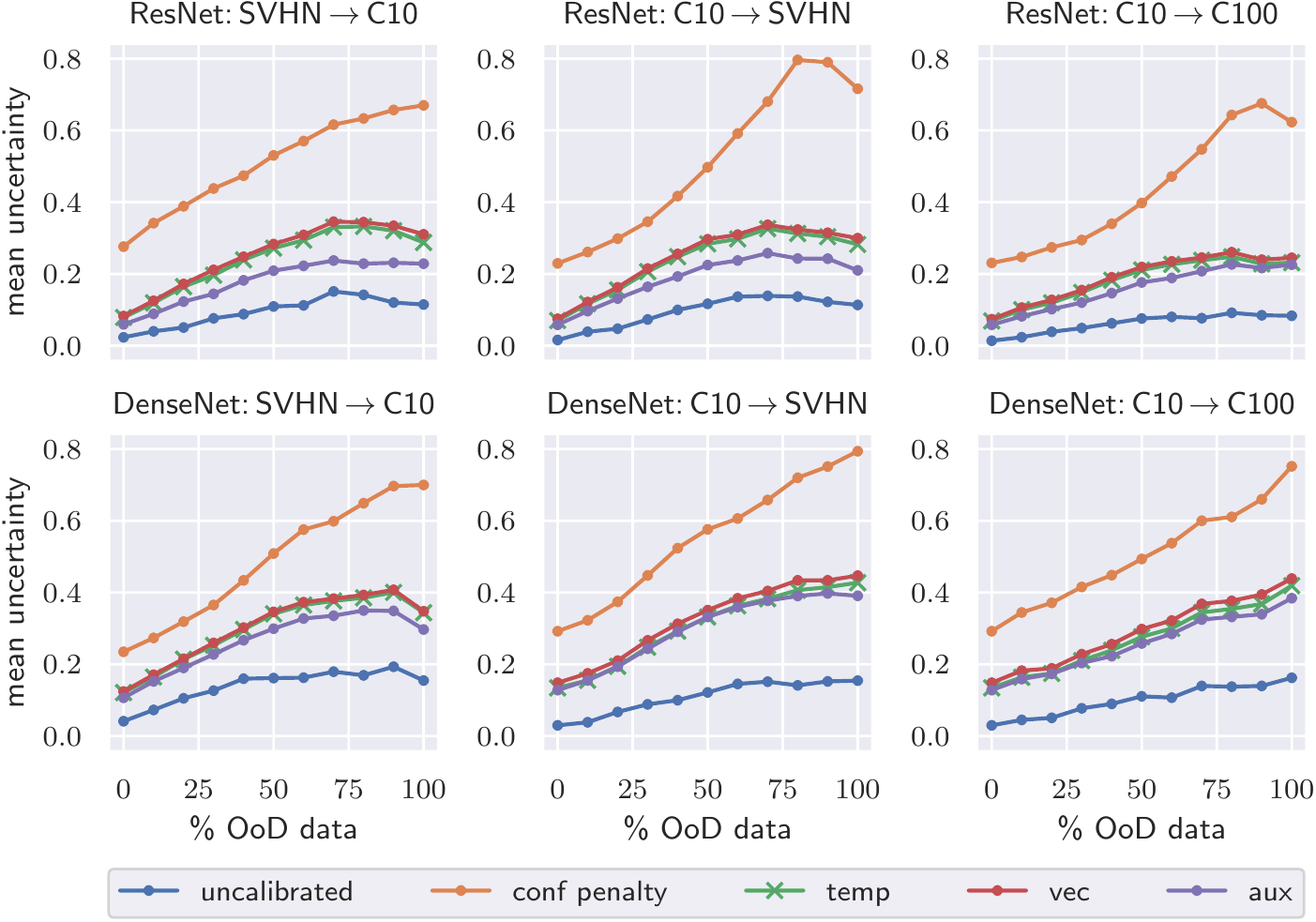}
    \end{tabular}
    \caption{(Left) The effect of the uncertainty threshold $ \mathcal{H}_{\mathrm{max}} $ on the test set error for the rejection of uncertain predictions. (Right) Test set results of out-of-distribution detection.}
    \label{fig:uncert_thresh}
\end{figure*}

In this section, the results of the above mentioned experimental setup are presented and discussed.

\subsection{Uncertainty Calibration}

Tab.\,\ref{tab:results} reports classwise UCE test set results and Fig.\,\ref{fig:reliability} shows reliability diagrams for the experimental setup described in the previous section.
All logit scaling methods considerably reduce miscalibration on CIFAR-10/100 by means of cECE and cUCE.
For the smaller networks on CIFAR-10 and SVHN, the more powerful aux scaling yields lowest cUCE.
On CIFAR-100, however, aux scaling increases miscalibration.
In this case, the auxiliary model $ \bm{R} $ has $ C=100 $ units in the hidden layer and easily overfits the calibration set (we observe calibration set accuracy of 100\,\%).
This results in worse calibration on the test set than the uncalibrated model.
A possible solution to this is adding regularization (e.\,g.\ early stopping or weight decay) during optimization of $ \bm{R} $.
If the model is already well-calibrated (e.\,g.\ for SVHN in our experiments), temperature scaling and vector scaling can slightly worsen calibration.
In this case, a larger calibration set is preferred or recalibration can be omitted at all.
Confidence penalty only slightly reduces miscalibration for larger models on CIFAR-100.
On all other configurations, it leads to worse calibration.
As hypothesized in §\,\ref{sec:related_work}, temperature scaling results in classwise calibrated uncertainty and is only marginally outperformed by the classwise logit scaling methods.
The reliability diagrams in Fig.\,\ref{fig:reliability} give additional insight and show, that calibrated uncertainty corresponds well with the model error.
It is worth noting that the likelihood in the Bayesian approach is generally better calibrated than the frequentist confidence.

\subsection{Rejection of Uncertain Predictions}

Fig.~\ref{fig:uncert_thresh} (left) shows the top-1 error as a function of decreasing $ \mathcal{H}_{\mathrm{max}} $.
For both uncalibrated and calibrated uncertainty, decreasing $ \mathcal{H}_{\mathrm{max}} $ reduces the top-1 error.
Again, we can observe the underestimation of uncalibrated uncertainty: $ \mathcal{H}_{\mathrm{max}} $ has little effect at first and few uncertain predictions are rejected.
Using calibrated uncertainty with temperature or vector scaling, the relationship is almost linear, allowing robust rejection of uncertain predictions.
Except for aux scaling on CIFAR-100, logit scaling is capable of reducing the top-1 error below 1\,\%.
Further, we observe that confidence penalty can lead to \emph{over}-estimation of uncertainty.

\subsection{Out-of-Distribution Detection}

Fig.\,\ref{fig:uncert_thresh} (right) shows the effect of calibrated uncertainty to OoD detection.
All calibration approaches are able to improve the detection of OoD data.
The benefit of calibration is most noticeable on ResNet (C10\,$\rightarrow$\,C100) and DenseNet (SVHN\,$\rightarrow$\,C10, C10\,$\rightarrow$\,SVHN), where the mean uncertainty stays almost constant for OoD data $ > 50\,\%$ and thus, robust OoD detection is only possible after calibration.
As in Fig.\,\ref{fig:uncert_thresh} (left), we can observe overestimation of uncertainty for confidence penalty.
In some cases (e.\,g.\ DenseNet SVHN\,$\rightarrow$\,C10), this causes a more robust OoD detection.
This is in contrast to the results presented in \cite{Lakshminarayanan2017}, where MC dropout uncertainty was not able to capture OoD data sufficiently.

\section{Conclusion}

In this paper, calibration of Bayesian model uncertainty is discussed.
We derive logit scaling as entropy maximization technique to recalibrate the uncertainty of deep models trained with Gaussian dropout.
Following commonly accepted metrics for calibration of confidence, we present the (classwise) expected uncertainty calibration error to quantify miscalibration of uncertainty.

Logit scaling calibrates uncertainty obtained by Monte Carlo Gaussian dropout with high effectiveness.
The experimental results show that better calibrated uncertainty allows more robust predictions and detection of out-of-distribution data; a key feature that is particularly important in safety-critical applications.
Logit scaling is easy to implement and more effective than confidence penalty during training.
Simple scaling methods are preferred over more complex methods, as they provide similar results and do not tend to overfit the calibration set.
Temperature scaling improves uncertainty estimation without affecting the accuracy of the model.
Vector and auxiliary scaling also improve calibration of uncertainty, but can have (positive or negative) influence on predictive accuracy.
By using entropy, the classwise uncertainty calibrated by vector and auxiliary scaling is not substantially better than that calibrated by temperature scaling.
Logit scaling calibrates not only the frequentist confidence but also the Bayesian uncertainty.

\section{Outlook}

Throughout this work, we used a fixed dropout rate $ p $ for Gaussian dropout.
In \cite{Gal2017}, the Concrete distribution was used as a continuous approximation to the discrete Bernoulli distribution in dropout, which allows optimizing $ p $ w.r.t.\ calibrated uncertainty.
Using Gaussian dropout as described above, we can also recalibrate models by optimizing $ p $ w.r.t.\ NLL on the calibration set, which scales $ \sigma $ to reduce underestimation of uncertainty.

In Bayesian active learning we want to train a model with the minimal number of expert queries from a pool of unlabeled data.
Calibrated uncertainty can further be useful to acquire the most uncertain samples from pool data to increase information efficiency \cite{Gal2017b}.

Additionally, pseudo-labels can be generated from the least uncertain predictions in semi-supervised learning.
However, there are many factors (e.\,g. network architecture, weight decay, dropout configuration) influencing the uncertainty in Bayesian deep learning that have not been discussed in this paper and are open to future work.

%

\small

\bibliography{literature}
\bibliographystyle{icml2020}

\appendix
\onecolumn

\section{Reviews}

This paper was submitted to \emph{International Conference on Machine Learning} (ICML) 2020 and rejected with the following scores:
\begin{itemize}
    \item Below the acceptance threshold, I would rather not see it at the conference.
    \item Borderline paper, but has merits that outweigh flaws.
    \item Borderline paper, but has merits that outweigh flaws.
    \item Borderline paper, but the flaws may outweigh the merits.
\end{itemize}
In the following, we disclose the anonymous reviews and our rebuttal.

\subsection{Meta-Review}

1. Please provide a meta-review for this paper that explains to both the program chairs and the authors the key positive and negative aspects of this submission. Because authors cannot see reviewer discussions, please also summarize any relevant points that can help improve the paper. Please be sure to make clear what your assessment of the pros/cons of this paper are, especially if your assessment is at odds with the overall reviewer scores. Please do not explicitly mention your recommendation in the meta-review (or you may have to edit it later).

The authors calibrate Gaussian dropout models and observe better calibrated uncertainty. After a discussion, the reviewers converged towards rejection being a more appropriate decision at this time. The reviewers agreed that the empirical evidence that model calibration is beneficial and that the analysis is sound. However, they generally felt that the novelty of the methods is limited and lacks justification.

8. I agree to keep the paper and supplementary materials (including code submissions and Latex source), and reviews confidential, and delete any submitted code at the end of the review cycle to comply with the confidentiality requirements.

Agreement accepted

9. I acknowledge that my meta-review accords with the ICML code of conduct (see https://icml.cc/public/CodeOfConduct).

Agreement accepted

\subsection{Review \#2}

\textbf{Questions}

1. Please summarize the main claim(s) of this paper in two or three sentences.

The authors apply the standard calibration techniques to Gaussian dropout models and observe better calibrated uncertainty.

2. Merits of the Paper. What would be the main benefits to the machine learning community if this paper were presented at the conference? Please list at least one.

The paper provides additional empirical evidence that model calibration is beneficial.

3. Please provide an overall evaluation for this submission.

Below the acceptance threshold, I would rather not see it at the conference.

4. Score Justification Beyond what you've written above as ''merits``, what were the major considerations that led you to your overall score for this paper?

The results of the paper are trivial. Temperature scaling as a well-known technique that improves the performance of basically all classification models. This particular paper applies it to Gaussian dropout networks.

5. Detailed Comments for Authors Please comment on the following, as relevant: - The significance and novelty of the paper's contributions. - The paper's potential impact on the field of machine learning. - The degree to which the paper substantiates its main claims. - Constructive criticism and feedback that could help improve the work or its presentation. - The degree to which the results in the paper are reproducible. - Missing references, presentation suggestions, and typos or grammar improvements.

I have read the author response. Some of my points have been addressed. I am willing to slightly increase my score, by I still think that the paper is below the acceptance threshold.

--

It is not clear to me why the authors focus on Gaussian dropout. Their main results, eq. 23-24, can be applied to any ensemble. Overall, the result is trivial: one can just take the predictive distribution of any model, be it a single neural network, a deep ensemble, or the result of MC dropout integration, and apply the temperature scaling, vector scaling or matrix scaling to this distribution. Moreover, the authors use Gaussian dropout as an approximation of binary dropout. Why do that when one can just start with Gaussian dropout? Moreover, since the authors mentioned the framework of variational inference, why not just stick with fully factorized Gaussian variational inference from the beginning? It has been a standard technique in Bayesian deep learning for years and does not require the extra steps going from binary dropout to its Bayesian interpretation, to its Gaussian approximation. This makes the paper much more confusing.

''the main contributions of our work are ... 3. first to apply logit scaling calibration to a Bayesian classifier obtained from MC dropout``
This has already been done by Ashukha et al 2020. They apply logit scaling to different kinds of ensembles, including 
Bayesian neural networks in general and beth MC dropout and FFG variational inference in particular.

The expected calibration error is a biased metric. Its bias depends on the model, so it cannot be used to compare the calibration of different models (Vaicenavicius et al 2019). The same holds for UCE. How is UCE different from other biased estimates of calibration error (ECE, TACE, SCE and others by Nixon et al 2019)? I am not convinced that this metric can provide any additional insights. Introducing more biased metrics is harmful to the community as it would make the further results on comparing different methods even less reliable. Moreover, there already are some calibration metrics that do not have such problems (Widmann et al 2019).

Nixon, Jeremy, et al. "Measuring calibration in deep learning." arXiv preprint arXiv:1904.01685 (2019).\\
Vaicenavicius, Juozas, et al. "Evaluating model calibration in classification." arXiv preprint arXiv:1902.06977 (2019).\\
Widmann, David, Fredrik Lindsten, and Dave Zachariah. "Calibration tests in multi-class classification: A unifying framework." Advances in Neural Information Processing Systems. 2019.\\
Ashukha, Arsenii, et al. "Pitfalls of In-Domain Uncertainty Estimation and Ensembling in Deep Learning." In International Conference on Learning Representations. 2020.

6. Please rate your expertise on the topic of this submission, picking the closest match.

I have published one or more papers in the narrow area of this submission.

7. Please rate your confidence in your evaluation of this paper, picking the closest match.

I tried to check the important points carefully. It is unlikely, though possible, that I missed something that could affect my ratings.

8. Datasets If this paper introduces a new dataset, which of the following norms are addressed? (For ICML 2020, lack of adherence is not grounds for rejection and should not affect your score; however, we have encouraged authors to follow these suggestions.)

This paper does not introduce a new dataset (skip the remainder of this question).

12. I agree to keep the paper and supplementary materials (including code submissions and Latex source) confidential, and delete any submitted code at the end of the review cycle to comply with the confidentiality requirements.

Agreement accepted

13. I acknowledge that my review accords with the ICML code of conduct (see https://icml.cc/public/CodeOfConduct).

Agreement accepted

\subsection{Review \#4}

\textbf{Questions}

1. Please summarize the main claim(s) of this paper in two or three sentences.

The authors propose a methodology for calibrating model uncertainty (measured as entropy of the marginal posterior predictive distribution) instead of the parameters of the (marginal) posterior predictive distribution (ECE). They introduce their approach in the context of MC Dropout, and demonstrate results on a set of experiments.

2. Merits of the Paper. What would be the main benefits to the machine learning community if this paper were presented at the conference? Please list at least one.

The authors propose the aforementioned methodology, and back it up with a set of empirical experiments.

3. Please provide an overall evaluation for this submission.
Borderline paper, but has merits that outweigh flaws.

4. Score Justification Beyond what you've written above as ''merits``, what were the major considerations that led you to your overall score for this paper?

The authors propose a method to calibrate the model uncertainty, but the indicated approach specifically calibrates the entropy of the marginal posterior predictive distribution, which contains both data and model uncertainty sources. Given that, I would have expected to see an experimental setup that compared the benefit of calibrating according to UCE vs. ECE. The listed experiments demonstrate that it is possible to apply calibration techniques developed for ECE to their proposed UCE, but the reader is left wondering whether UCE provides a marked improvement. The rejection experiments are useful, but it would have been good to compare the results to the alternative of thresholding on the max predicted probability (i.e., Hendrycks et al., 2017). I agree with the authors in the motivation for using model uncertainty, but I still think the paper would benefit from the comparison.

Addendum:\\
Thank you to the authors for the rebuttal! Given the noted inclusion of UCE vs. ECE experiments, comparison to max predicted probability, added discussion around Nixon et al. 2019, and updated text re: predictive entropy containing both data \& model uncertainty, I have increased my score.

5. Detailed Comments for Authors Please comment on the following, as relevant: - The significance and novelty of the paper's contributions. - The paper's potential impact on the field of machine learning. - The degree to which the paper substantiates its main claims. - Constructive criticism and feedback that could help improve the work or its presentation. - The degree to which the results in the paper are reproducible. - Missing references, presentation suggestions, and typos or grammar improvements.

Significance: Considering model uncertainty and the extent to which it is calibrated is well-motivated, as is the usage of it for making rejections in order to improve performance. The experiments indicate that there is promise in both calibrating measures that incorporate model uncertainty. However, the experiments do not directly demonstrate the benefit over existing baselines using ECE and the parameters of the (marginal) predictive distribution. One other issue is that the entropy of the marginal posterior predictive distribution is a measure of both data uncertainty and model uncertainty.

Novelty: To the best of the reviewer's knowledge, a calibration metric for predictive entropy has not been introduced before.

Presentation/clarity:

- p. 1, line 21: "considerably reduce" -> ''considerably reduces``\\
- p. 2, line 83, left: define ECE and cite Naeini et al., 2015.

6. Please rate your expertise on the topic of this submission, picking the closest match.

I have published one or more papers in the narrow area of this submission.

7. Please rate your confidence in your evaluation of this paper, picking the closest match.

I tried to check the important points carefully. It is unlikely, though possible, that I missed something that could affect my ratings.

8. Datasets If this paper introduces a new dataset, which of the following norms are addressed? (For ICML 2020, lack of adherence is not grounds for rejection and should not affect your score; however, we have encouraged authors to follow these suggestions.)

This paper does not introduce a new dataset (skip the remainder of this question).

12. I agree to keep the paper and supplementary materials (including code submissions and Latex source) confidential, and delete any submitted code at the end of the review cycle to comply with the confidentiality requirements.

Agreement accepted

13. I acknowledge that my review accords with the ICML code of conduct (see https://icml.cc/public/CodeOfConduct).

Agreement accepted

\subsection{Review \#5}

\textbf{Questions}

1. Please summarize the main claim(s) of this paper in two or three sentences.

The main claims are a new metric for uncertainty calibration and the introduction of logit scaling with Gaussian MC Dropout. The logit scaling with MC dropout is analyzed empirically.

2. Merits of the Paper. What would be the main benefits to the machine learning community if this paper were presented at the conference? Please list at least one.

Calibration and Bayesian approaches are often seen as two, different approaches for obtaining better calibrated predictions. This paper shows that calibration is also beneficial for Bayesian DNNs. Furthermore, uncertainty in deep learning is a highly relevant topic, as it is substantial for real world deep learning in safety critical environments. Additional insight is always welcome for advancing the field.

3. Please provide an overall evaluation for this submission.

Borderline paper, but has merits that outweigh flaws.

4. Score Justification Beyond what you've written above as "merits", what were the major considerations that led you to your overall score for this paper?

The paper is well written and the analysis is sound. It can still be improved, but due to the importance of the topic and the works quality, i deem it over the acceptance threshold. The novelty of the methods is limited, but additional insight is sufficient for advancing a field.

5. Detailed Comments for Authors Please comment on the following, as relevant: - The significance and novelty of the paper's contributions. - The paper's potential impact on the field of machine learning. - The degree to which the paper substantiates its main claims. - Constructive criticism and feedback that could help improve the work or its presentation. - The degree to which the results in the paper are reproducible. - Missing references, presentation suggestions, and typos or grammar improvements.

The novel methods (uncertainty calibration metric and logit scaling for gaussian dropout) are straight forward applications of known principles and ideas. However, the paper is still somewhat significant due to the novel insight presented by the authors. Especially the combination of Bayesian approximations and calibration is relevant, as it was recently shown that Bayesian methods are not always leading to better calibrated predictions. The paper can trigger additional research into the application of calibration methods for Bayesian approximations, which is especially interesting when considering that Bayesian methods are still very expensive and calibrated Bayesian methods may offer a way to mitigate the flaws of cheaper posterior approximations.

The claims of the paper are sufficiently substantiated. Approaches and equations are well explained and understandable. However, as the paper is mostly depending on the results and analysis, this section should be extended. Some possible improvements are: Comparison with Dirichlet calibration, better comparison with frequentist results (e.g. cECE) and analysis of class distribution changes (within the same dataset).
The results are likely reproducible, due the available code and the use of standard DNN architectures.

6. Please rate your expertise on the topic of this submission, picking the closest match.

I have seen talks or skimmed a few papers on this topic, and have not published in this area.

7. Please rate your confidence in your evaluation of this paper, picking the closest match.

I am willing to defend my evaluation, but it is fairly likely that I missed some details, didn't understand some central points, or can't be sure about the novelty of the work.

12. I agree to keep the paper and supplementary materials (including code submissions and Latex source) confidential, and delete any submitted code at the end of the review cycle to comply with the confidentiality requirements.

Agreement accepted

13. I acknowledge that my review accords with the ICML code of conduct (see https://icml.cc/public/CodeOfConduct).

Agreement accepted

\subsection{Review \#6}

\textbf{Questions}

1. Please summarize the main claim(s) of this paper in two or three sentences.

The authors study the problem of calibration of uncertainty inspired by calibration of confidence. Specifically, the authors modify several existing calibration methods to do calibration of uncertainty for Gaussian dropout. The proposed methods are tested on standard calibration tasks in comparison with the corresponding calibration of confidence methods.

2. Merits of the Paper. What would be the main benefits to the machine learning community if this paper were presented at the conference? Please list at least one.

The idea of calibration of uncertainty is interesting and reasonable. As far as I understand, this is the first work to give an attempt.

3. Please provide an overall evaluation for this submission.

Borderline paper, but the flaws may outweigh the merits.

4. Score Justification Beyond what you've written above as "merits", what were the major considerations that led you to your overall score for this paper?

Although it is interesting to see a paper attempting calibration of uncertainty, the method is very handwavy and lack of justification.

5. Detailed Comments for Authors Please comment on the following, as relevant: - The significance and novelty of the paper's contributions. - The paper's potential impact on the field of machine learning. - The degree to which the paper substantiates its main claims. - Constructive criticism and feedback that could help improve the work or its presentation. - The degree to which the results in the paper are reproducible. - Missing references, presentation suggestions, and typos or grammar improvements.

Compared to previous methods, the only difference is replacing the confidence probability by uncertainty which is measured by normalized entropy. The use of normalized entropy as an uncertainty metric and the definition of the perfect calibration of uncertainty still need justification. The authors did not provide a clear connection of normalized entropy and uncertainty as well as a connection between normalized entropy and top-1 error. Therefore, the basis of all the proposed methods in the paper seems very handwavy.

For the experiments, the authors seem to only compare with ECE in the first experiment. It will be better to report the ECE results on the other experiments as well. I’m curious if calibrated MC dropout is better than a calibrated point estimate. From the results of the first experiment, it did not seem to be true. 

Update: Thank the authors for the clarification. However, without seeing the new results, the concerns about experiments remain. Thus I keep the original score.

6. Please rate your expertise on the topic of this submission, picking the closest match.

I have seen talks or skimmed a few papers on this topic, and have not published in this area.

7. Please rate your confidence in your evaluation of this paper, picking the closest match.

I am willing to defend my evaluation, but it is fairly likely that I missed some details, didn't understand some central points, or can't be sure about the novelty of the work.

8. Datasets If this paper introduces a new dataset, which of the following norms are addressed? (For ICML 2020, lack of adherence is not grounds for rejection and should not affect your score; however, we have encouraged authors to follow these suggestions.)

This paper does not introduce a new dataset (skip the remainder of this question).

12. I agree to keep the paper and supplementary materials (including code submissions and Latex source) confidential, and delete any submitted code at the end of the review cycle to comply with the confidentiality requirements.

Agreement accepted

13. I acknowledge that my review accords with the ICML code of conduct (see https://icml.cc/public/CodeOfConduct).

Agreement accepted

\subsection{Rebuttal}

1. Author Response to Reviewers Please use this space to respond to any questions raised by reviewers, or to clarify any misconceptions. Please do not include any links to external material, nor include ''late-breaking`` results that are not responsive to reviewer concerns. We request that you understand that this year is especially difficult for many people, and to be considerate in your response.

We thank the reviewers for their valuable feedback. It allows us to improve our paper substantially.

We acknowledge Reviewer \#2's references to Ashukha et al., (2020) and other highly relevant work and will update our literature review accordingly.
Reviewer \#2's main concern seems to be the disadvantages of ECE-like calibration metrics.
After carefully reading the suggested literature (Widmann et al, 2019; Ashukha et al., 2020; Nixon et al., 2019), two major concerns with recent calibration metrics are raised, which do not apply to UCE:
1. Non-applicability to multi-class classification:
In contrast to ECE, UCE considers all class predictions by using the predictive entropy as uncertainty metric. We already addressed that in our manuscript and compare to classwise ECE as suggested by Kull et al., (2019).
2. ''ECE-like scores are minimized by a model with constant uniform predictions`` (Ashukha et al., 2020; and analogously Nixon et al., 2019):
This also does not apply to the UCE metric as uniform predictions would result in high entropy. Consider the following example: Binary classification with balanced class frequencies and a model with constant uniform predictions. This would result in ECE=0\%, but UCE=50\%.

UCE suffers from fixed bin sizes (Nixon et al., 2019), which we will discuss appropriately in our conclusion. This could easily be fixed by combining UCE with adaptive binning from ACE/TACE.
We do not believe that the proposed UCE metric is harmful to the community as it does not have the major disadvantages compared to other ECE-like metrics.
UCE is a useful metric and can give valuable insights into the calibration of uncertainty.

We focus on Gaussian dropout as we have derived our approach from the MC dropout framework for uncertainty estimation. We will adjust this section and refer to fully factorized Gaussian variational inference to reduce the reader's confusion.

We thank reviewer \#2 for pointing out that temperature scaling was recently applied to MC dropout by Ashukha et al., (2020). We further extend their work by applying more complex logit scaling calibration to a Bayesian classifier obtained from MC dropout. Our work therefore provides additional insights into calibration of Bayesian neural nets. Our results suggest that the more complex calibration methods (like class-wise calibration) is advantageous compared to only temperature scaling (see bold values in Tab. 1).

Based on feedback from reviewers \#4 and \#6, we extended our experiments to emphasize the benefits of calibration according to UCE vs. ECE.
We now also compare the rejection and OoD detection experiments to thresholding on the max predicted probability (i.e., Hendrycks et al., 2017).
We added additional figures and corresponding text passages to the results section of the manuscript.

Based on the comment of Reviewer \#6 we realized the lack of a clear connection between normalized entropy and uncertainty/top-1 error.
The use of predictive entropy to measure predictive uncertainty in classification is well motivated in Gal, (2016) pp. 51--54.
Normalization was introduced to restrict the values to [0, 1] independent of the number of classes C.
Normalization is not essential for calibration but gives a more "intuitive" interpretation of the uncertainty values themselves.
When all entries of the probability vector are predicted with equal probability, normalized entropy equals to 1.0 and we expect the prediction to be false (i.e. the expectation of the top-1 error to be 1.0).
We added a more detailed explanation on the use of normalized entropy to the manuscript.

Reviewer \#4 mentioned that "the entropy of the marginal posterior predictive distribution is a measure of both data uncertainty and model uncertainty".
Classification models trained by minimizing NLL (i.e. cross-entropy) already capture a data-dependent uncertainty.
Therefore, the predictive entropy both contains data and model uncertainty.
We added a sentence for clarification and changed the manuscript accordingly.

We hope that our revisions meet the expectations of the reviewers. The comments have greatly helped us to increase the quality of our work. We thank the reviewers for their valuable time.

Nixon, J. et al. "Measuring calibration in deep learning." arXiv preprint arXiv:1904.01685 (2019).\\
Widmann, D. et al. "Calibration tests in multi-class classification: A unifying framework." Advances in Neural Information Processing Systems. 2019.\\
Ashukha, A. et al. "Pitfalls of In-Domain Uncertainty Estimation and Ensembling in Deep Learning." In International Conference on Learning Representations. 2020.\\
Gal, Y. Uncertainty in Deep Learning. PhD thesis, Department of Engineering, University of Cambridge, 2016

3. I certify that this author response conforms to the ICML Code of Conduct (https://www.icml.cc/public/CodeOfConduct)

Agreement accepted

\end{document}